\documentclass[conference]{IEEEtran}
\IEEEoverridecommandlockouts                              
\usepackage{bm}

\usepackage{amsmath,amssymb,amsfonts}
\usepackage{algorithmic}
\usepackage{graphicx}
\usepackage{textcomp}
\usepackage{xcolor}
\usepackage{fancyhdr}
\usepackage{url} % not crucial - just usted below for the URL
\def\BibTeX{{\rm B\kern-.05em{\sc i\kern-.025em b}\kern-.08em
    T\kern-.1667em\lower.7ex\hbox{E}\kern-.125emX}}
\usepackage{cite}
\usepackage{hyperref}
\title{\LARGE \bf
A Comprehensive Survey of PID and Pure Pursuit Control Algorithms for Autonomous Vehicle Navigation
}

\author{\IEEEauthorblockN{ Harshit Jain$^*$}
\IEEEauthorblockA{\textit{Department of Computer Engineering} \\
\textit{K.J. Somaiya College Of Engineering} \\ 
\textit{Vidyavihar, Mumbai, India}\\
\textit{harshit.nj@somaiya.edu}\\
}
\and
\IEEEauthorblockN{Priyal Babel$^*$}
\IEEEauthorblockA{\textit{Department of Computer Engineering} \\
\textit{K.J. Somaiya College Of Engineering} \\ 
\textit{Vidyavihar, Mumbai, India}\\
\textit{priyal.babel@somaiya.edu}\\
}
}

\begin{document}

\maketitle
\thispagestyle{empty}
\pagestyle{empty}

\def\thefootnote{*}\footnotetext{These authors contributed equally to this work}\def\thefootnote{\arabic{footnote}}

%%%%%%%%%%%%%%%%%%%%%%%%%%%%%%%%%%%%%%%%%%%%%%%%%%%%%%%%%%%%%%%%%%%%%%%%%%%%%%%%
\begin{abstract}

The autonomous driving industry is experiencing unprecedented growth, driven by rapid advancements in technology and increasing demand for safer, more efficient transportation. At the heart of this revolution are two critical factors: lateral and longitudinal controls, which together enable vehicles to track complex environments with high accuracy and minimal errors. This paper provides a detailed overview of two of the field's most commonly used and stable control algorithms: proportional-integral-derivative (PID) and pure pursuit. These algorithms have proved useful in solving the issues of lateral (steering) and longitudinal (speed and distance) control in autonomous vehicles. This survey aims to provide researchers, engineers, and industry professionals with an in depth understanding of these fundamental control algorithms, their current applications, and their potential to shape the future of autonomous driving technology.

\end{abstract}


%%%%%%%%%%%%%%%%%%%%%%%%%%%%%%%%%%%%%%%%%%%%%%%%%%%%%%%%%%%%%%%%%%%%%%%%%%%%%%%%
\section{INTRODUCTION}
A driverless vehicle, also known as an autonomous vehicle (AV) or self-driving car, is a vehicle capable of sensing its environment and operating without human involvement. These vehicles combine a variety of sensors to perceive their surroundings, advanced control systems to interpret sensory information, and complex algorithms to identify appropriate navigation paths, obstacles, and relevant signals. The primary goal of this technology is to reduce human error in driving, thereby increasing safety, efficiency, and accessibility in transportation. 

An autonomous driving system is fundamentally built upon the learning, understanding, refinement, and optimization of three critical driving processes: environment perception, path planning, and control execution. These components work in concert to replicate and enhance the core elements of human driving, aiming to create a safer and more efficient transportation solution.

The perception module is responsible for understanding the vehicle's environment. It processes data from various sensors such as LiDAR (Light Detection and Ranging), radar, camera and GPS (Global Positioning System). This module performs tasks such as object detection, categorization, and tracking, as well as localization of the vehicle within its environment.

Once the environment is perceived, the path planning module determines the optimal route for the vehicle to follow. This involves both global path planning (determining the overall route from start to destination) and local path planning (navigating immediate obstacles and traffic conditions). The module must consider factors such as road rules, safety, efficiency, and passenger comfort.

The control module is responsible for executing the planned path by manipulating the vehicle's actuators. This module translates high-level commands from the path planning module into specific instructions for the vehicle's steering, acceleration, and braking systems. It ensures that the vehicle follows the intended path while maintaining stability and smoothness of motion.

\begin{figure*}
\centering
\includegraphics[width=0.9\textwidth]{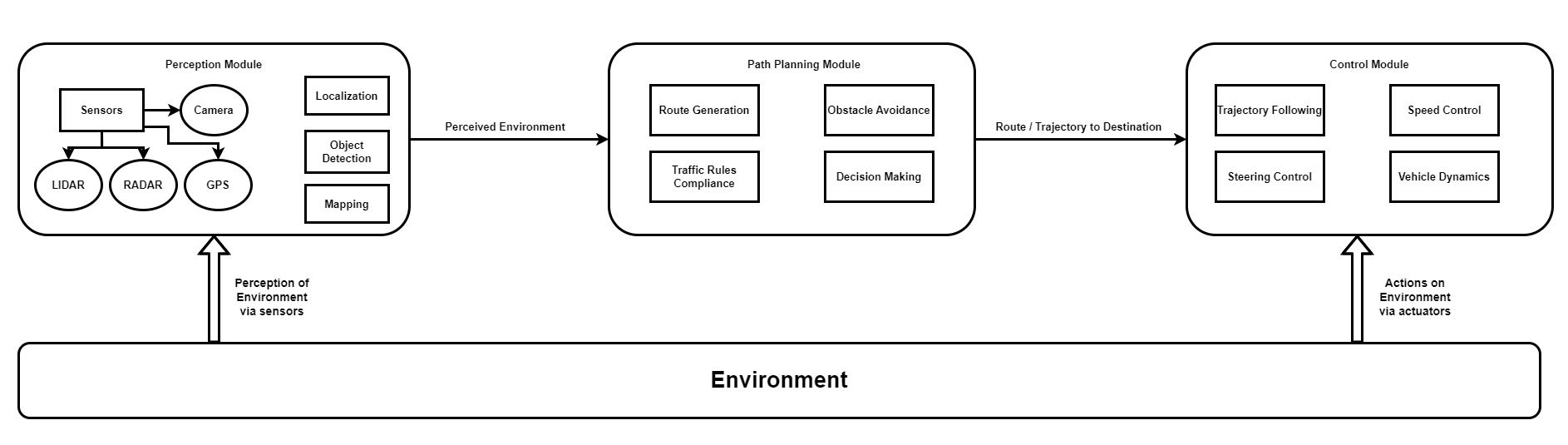}
\caption{Autonomous System}
\label{autonomous-system}
\end{figure*}

This system operates through a continuous feedback loop. As the vehicle moves, its location relative to its surroundings shifts, creating changes in the environment. The sensors detect these changes quickly and restart the cycle. This continuous loop allows the vehicle to constantly adapt to its changing surroundings, making real-time modifications to safely reach its destination. The cycle is depicted the in figure \ref{autonomous-system}.

As we can see, the control module in autonomous vehicles plays a critical role in translating decisions into actions that directly impact the environment, making its accuracy and reliability of utmost importance. A single faulty signal from this module could cause serious effects, highlighting the need for strong and dependable control algorithms. This paper presents a detailed analysis of two commonly used and stable control algorithms: proportional-integral-derivative (PID) and pure pursuit. We have examined several studies in this area, assessing their strengths and weaknesses. By examining these algorithms' applications in autonomous vehicles, we aim to provide insights into their effectiveness, adaptability, and potential areas for improvement. The paper concludes by exploring emerging trends and future developments in autonomous vehicle control systems, offering a forward-looking perspective on advancements that could enhance safety, efficiency, and performance in this rapidly evolving field.

\section{Control Algorithms}
Before moving onto the research papers, let's first understand the basics of our two algorithms. 

\subsection{PID}

\begin{figure}[htp]
\centerline{\includegraphics[width=0.52\textwidth]{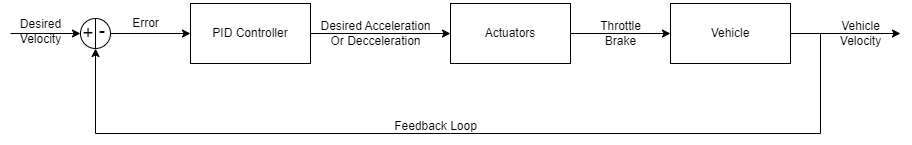}}
\caption{PID Feedback Loop}
\label{PID Feedback Loop}
\end{figure}

The Proportional-Integral-Derivative (PID) algorithm is a basic control mechanism used in various applications, including the longitudinal control of autonomous vehicles. The algorithm takes the vehicle's current speed and tries to match it with the target speed (output from the path planning module) by adjusting the throttle and brake commands. The difference between the speeds is the error that needs to be zero in ideal scenarios. PID functions as a feedback loop, and with each iteration, the algorithm attempts to minimize this error. The feedback loop is depicted in the figure \ref{PID Feedback Loop}.

The algorithm comprises of three key components:

Proportional term: Directly proportional to the current error
Integral term: Proportional to the integral of the error over time
Derivative term: Proportional to the derivative of the error

The PID controller in mathematical terms is expressed as:

\[ u(t) = K_p \cdot e(t) + K_i \int e(t) dt + K_d \frac{de(t)}{dt}\]

Where:
u(t) is the control signal
e(t) is the error (difference between desired and current speed)
Kp, Ki, and Kd are the proportional, integral, and derivative gains, respectively

The efficiency of a PID controller is heavily dependent on the proper tuning of its gains (Kp, Ki, and Kd). There are many algorithms for PID tuning, the most popular of which is the Ziegler-Nichols method. When adjusting PID controllers, many key performance criteria are taken into consideration:

\noindent Rise time: The time required to reach 90\% of the reference value \\ 
Overshoot: The output exceeds the reference by a maximum percentage of \\
Settling time: The time taken to settle within ±5\% of the reference value \\
Steady-state error: The error between the output and the reference at steady-state \\

Adjusting each gain affects the system's response in specific ways which are mentioned in the table \ref{pid-tuning}: 

\begin{table}[htbp]
\centering
\caption{Tuning PID}
\begin{tabular}{|c|c|c|}\hline
\textbf{Increasing \(K_p\)} & \textbf{Increasing \(K_d\)} & \textbf{Increasing \(K_i\)} \\\hline
 Strengthens reaction & Decreases & Eliminates steady \\ to errors  & overshoot & state errors\\\hline
 Decreases & Reduces & May increase \\ rise time  & oscillations & oscillations \\\hline
 May increase & Potentially decreases &  \\ overshoot  & settling time &  \\\hline
\end{tabular}
\label{pid-tuning}
\end{table}

It is important to note that these gains interact with each other, and their combined effects must be considered to achieve optimal closed-loop performance.

\subsection{Pure Pursuit}
Lateral control is an important component in autonomous vehicle navigation because it ensures that a vehicle accurately follows a desired path. The fundamental goal of lateral control is to generate appropriate steering inputs that allow the vehicle to align and maintain its position on the desired trajectory. This trajectory is given by the path planning module, which takes into account parameters such as road geometry, obstacles, and destination. Effective lateral control is essential for safe and efficient autonomous navigation since it directly affects the vehicle's ability to remain in its designated lane, handle curves, and avoid obstacles.

To achieve accurate lateral control, two key error metrics are constantly evaluated and reduced: heading error and crosstrack error. The heading error (\(\psi_e\)) is defined as the difference between the path direction (\(\psi_p\)) and the vehicle heading (\(\psi_v\)) at a given point on the path, expressed as \(\psi_e\) = \(\psi_p\) - \(\psi_v\). This error quantifies how well the vehicle is aligned with the desired path direction. The crosstrack error (\(e_c\)), sometimes called offset error, is the perpendicular distance between the vehicle's reference point and the nearest point on the desired path. It can be calculated as (\(e_c\)) = \( || p_v - p_p ||\), where (\(p_v\)) is the vehicle position and (\(p_p\)) is the nearest point on the path. The crosstrack error indicates how near the vehicle is to the target point along the path. For optimal path tracking, these errors should approach to zero, ensuring that the vehicle is properly aligned with and positioned on the desired trajectory.

\begin{figure}
\centerline{\includegraphics[width=0.4\textwidth]{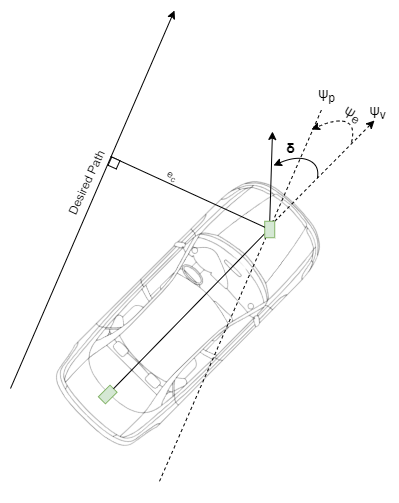}}
\caption{Lateral Errors}
\label{Lateral Errors}
\end{figure}

\noindent Where: \\ 
\(e_c\) - Crosstrack error \\
\(\psi_e\) - Heading error \\
\(\psi_p\) -  Path direction \\
\(\psi_v\) - Vehicle heading \\
\(\delta\) - Steering angle \\

The pure pursuit method stands on the concept that a reference point is placed on the desired path at a fixed distance ahead of the vehicle. The steering command must reach this point by providing a constant steering angle. As the vehicle advances and turns to approach the point, the point moves ahead, resulting in a new steering angle calculation. This cycle will continue till the car reaches its destination. The look ahead distance (\(l_d\)) refers to the distance between the target point on the path and the vehicle's reference point. 

The pure pursuit approach is mathematically based on the bicycle model of vehicle kinematics, with the center of the rear axle acting as the vehicle's reference point. The key parameter in this model is \(\alpha\), the angle between the vehicle's longitudinal axis and the line connecting the rear axle center to the reference point on the path. Using this geometry, the steering angle \(\delta\) can be calculated as:

\[\delta = \tan^{-1}\left( \frac{2L\sin\alpha}{l_d}\right) \] 

\noindent \(\delta\) - steering angle \\
L - wheelbase of the vehicle \\
\(l_d\) - lookahead distance \\
\(\alpha\) - angle between the vehicle's longitudinal axis and the line connecting the rear axle center to the reference point on the path \\ 

This equation assumes small steering angles and ignores slip conditions. The look-ahead distance, \(l_d\), plays an important role in defining the controller's behavior, with shorter distances giving more aggressive steering reactions and longer distances producing smoother, more gradual changes.

The pure pursuit controller is a straightforward control method that neglects the dynamic forces acting on vehicles and presumes the wheels experience no slip. Additionally, if it is calibrated for low-speed operation, it becomes excessively aggressive and potentially hazardous at higher speeds. An enhancement to this controller is to adjust the look-ahead distance \(l_d\) according to the vehicle's speed.

\[l_d = K_{dd} \cdot V_f \]

So the steering angle changed as:

\[ \delta = \tan^{-1} \left( \frac{2L \sin \alpha}{K_{dd}v_f} \right)\]

\section{Survey}
\subsection{Development of Lateral Control System for Autonomous Vehicle Based on Adaptive Pure Pursuit Algorithm }
The researchers of \cite{b1} proposed an improved lateral control system for autonomous vehicles based on an adaptive pure pursuit algorithm. While effective, the traditional pure pursuit approach tends to cut corners at high speeds and on curved courses since it relies on a velocity-dependent look-ahead distance. To overcome this issue, the researchers developed an adaptive approach that combines the traditional pure pursuit method with additional control elements.
The key improvement lies in the incorporation of a PI (Proportional-Integral) control theory applied to the lateral offset. This enhancement aims to reduce tracking errors, particularly on curved paths. By applying this control to the lateral offset, with a constant proportional gain and a curvature-dependent integral gain, the researchers created a more responsive system. The total desired steering angle in their improved method is calculated as the sum of two components: the desired angle based on the look-ahead distance (from the traditional pure pursuit approach) 

\[ \delta_{l_d}(t) = \tan^{-1}\left(\frac{2L \sin(\alpha(t))}{l_d}\right)\]

and the desired angle based on the lateral offset (derived from their PI control enhancement)

\[ \delta_{e_y}(t) = Re_y(t) + Q(\kappa(t)) \int e_y \cdot dt \]

R is proportional gain and Q(K(t)) is integral gain

\[ \delta_t = \delta_{l_d} + \delta_{e_y}\]
Hence, allowing for more precise path tracking, particularly on curved roads.
 
The researchers tested their improved algorithm on a 1.4 km test track with various curved path segments. The results demonstrated notable improvements in path tracking accuracy compared to the original pure pursuit method, with error reductions of up to 1.5 meters on some curves. While the enhanced algorithm showed similar performance to the original on low-curvature paths, its overall improvement in handling curves represents a significant step forward in autonomous vehicle control.

\subsection{Pure Pursuit Revisited: Field Testing of Autonomous Vehicles in Urban Areas}

The researchers in \cite{b2} implemented the Pure Pursuit path following algorithm for autonomous vehicles and identified several problems with the traditional approach through simulations and field tests. These issues included vehicles meandering off the intended path, unstable steering control, and cutting the insides of tight corners. To address these problems, they focused on tuning two key aspects of the algorithm: the lookahead distance parameter and incorporating linear interpolation.

The lookahead distance was made dynamic and velocity-dependent. They defined the lookahead distance as the distance the vehicle would travel in a certain number of seconds at the current speed. 
\[l_d = v_cx\]
Here, \(l_d\) [m] is the lookahead Distance, \(v_c\) [m/s] is the current linear velocity and x [s] is a magnification parameter. By tuning this parameter, they could balance between smooth path following at higher speeds and accurate cornering at lower speeds. Simulation results showed that smaller lookahead distances allowed tighter cornering but could lead to oscillations, while larger values resulted in smoother trajectories but caused the vehicle to cut inside curves.

The meandering or unstable steering control was due to the angular velocity which was calculated from the discrete waypoints on the given path. To tackle this, the researchers introduced a linear interpolation approach to generate a continuous target point, rather than just selecting the nearest discrete waypoint outside the lookahead distance. This modification involved finding the intersection between the search circle (defined by the lookahead distance) and the line connecting the next two waypoints on the path. Interpolating between distinct waypoints allowed them to generate more compact points leading to a smoother and more continuous steering commands. The graphs of the resulting angular velocity showed that linear interpolation eliminated the discrete steps seen in the traditional approach, leading to more stable steering control and improved path following performance.

\subsection{Coordinated PSO-PID based longitudinal control with LPV-MPC based lateral control for autonomous vehicles}

The authors of \cite{b3} provided an optimized PID controller for longitudinal control of autonomous cars. To handle the task of speed tracking based on nonlinear longitudinal dynamics, the researchers developed an enhanced PSO algorithm to optimize the PID gains (Kp, Ki, Kd). The PSO algorithm works by simulating a swarm of particles, each representing a potential solution (i.e., a set of PID gains). These particles move through the solution space, updating their positions based on their own best known position and the swarm's global best position. The algorithm is given by:

\begin{multline*}    
\begin{cases}
v_i(k+1) = \omega v_i(k) + c_1r_1(Pb_i(k) - x_i(k))\\ \qquad \qquad \qquad+ c_2r_2(Gb(k) - x_i(k)) \\
x_i(k+1) = x_i(k) + v_i(k+1)
\end{cases}
\end{multline*}

where \(\omega\), \(c_1\) and \(c_2\) are known as inertia weight, cognitive and social accelerations respectively, and \(r_{1,2}\) \(\in\) [0, 1] are random constants. Gb and Pb are the global best position of the whole swarm and the local best position in a single swarm generation. The effectiveness of each solution is evaluated using the Mean Squared Error (MSE) of the speed tracking performance. Authors have improved this PSO version by using dynamic inertia weight, social, and cognitive acceleration parameters that enhance the overall search capabilities of the algorithm. 

The results of the PSO-PID controller for longitudinal control are encouraging. In a double lane-change maneuver test with varying velocity between 50 and 65 km/h and exposure to wind disturbances, the improved PID controller accomplished smooth speed tracking with adequate accuracy, resulting in a Mean Squared Error (MSE) of 0.0213. For a more complicated general trajectory with multiple bends and a varying velocity profile, the PSO optimization gave a different set of PID gains, but the controller still performed well with an MSE of 0.0187. The control signals in both scenarios were smooth, indicating comfortable ride quality. These results suggest that the PSO-PID technique for longitudinal control can handle a variety of driving circumstances and disturbances while maintaining accurate speed tracking.

\subsection{Lateral Control of an Autonomous Vehicle Based on Pure Pursuit Algorithm}

The authors in \cite{b4} presented an implementation of a pure pursuit algorithm for lateral control of an autonomous vehicle. The authors used both linear and non-linear kinematic bicycle models to simulate the vehicle's response to steering inputs in OCTAVE. Their implementation of the pure pursuit algorithm is that of the traditional algorithm which works by selecting a goal point on the desired path that is a certain "look-ahead distance" ahead of the vehicle. It then calculates the required steering angle to follow a circular arc that passes through both the vehicle's current position and the goal point.

The authors derive the steering angle equation using the geometry of the pure pursuit method. They use the law of sines to relate the look-ahead distance, the radius of curvature, and the angle between the vehicle's heading and the look-ahead vector. The resulting equation calculates the required steering angle as a function of the vehicle's wheelbase length, the look-ahead distance, and the angle to the goal point.

\[ \delta(t) = \tan^{-1}\left(\frac{2L \sin(\alpha(t))}{l_d}\right)\]

The authors tested their pure pursuit implementation on various path shapes including straight lines, circular paths, elliptical paths, and curved paths. They simulated the vehicle's response using both linear and non-linear kinematic models at different speeds (e.g., 8 m/s for straight lines, 5 m/s for circular paths, 9 m/s for elliptical paths). Their findings show that the algorithm effectively follows the desired paths with minimal cross-track errors. For example, in circular route tracking, they reported an average cross-track error of 0.88311 m. They also observed that the non-linear model needed somewhat more steering inputs than the linear model, which they ascribe to the inclusion of non-linear parameters in that model.

\subsection{Path Tracking Based on Improved Pure Pursuit Model and PID}

To overcome the problems faced with the traditional pure pursuit algorithm, the researchers in \cite{b5} tweaked the original algorithm and combined it with a PID (Proportional-Integral-Derivative) controller. They modified the pure pursuit model to adjust the look-ahead distance dynamically based on the vehicle’s speed and the curvature of the path. The updated look ahead distance is given by

\[l_c = l_0 + k_1v+k_2w\]

\(l_0\) represents the basic value of the look ahead distance, \(k_1\) and \(k_2\) is are constant values that represent the correlation coefficients of speed v and curvature of path w. \(k_2\) is a negative value and it meant that the greater the correlation coefficient of path’s curvature, the smaller the look ahead distance. This allowed the algorithm to be more responsive to changes in the path and the vehicle state. 

They then combined this improved pure pursuit model with a PID controller. The PID controller takes the error between the desired and actual vehicle states (like position and orientation) and calculates a correction term. This correction is applied to the steering angle originally calculated by the pure pursuit algorithm. By combining these two approaches, the researchers created a hybrid algorithm that could more accurately calculate the required steering angle.  

The results showed significant improvements in path tracking performance. In straight-line tests, the combined pure pursuit-PID method reduced the mean tracking error by 15.13\% compared to pure pursuit alone. The hybrid algorithm performed better on curved courses, especially in maintaining accuracy during turns. Overall, the enhanced algorithm resulted in more stable and precise vehicle control across different paths and speeds.

\subsection{Vehicle Speed Control for Autonomous Vehicle using PID Controller}

The researchers of \cite{b6} implemented a Proportional-Integral-Derivative (PID) controller to regulate the speed of an autonomous vehicle. The system used a Raspberry Pi 3B+ as the microcontroller, programmed in Python, to control a linear motor attached to the vehicle. An incremental encoder was used as feedback to calculate the vehicle's actual speed. The PID controller was developed to maintain the vehicle's speed at a given target by continuously adjusting the motor's output based on the error between the desired and actual speed. 

The researchers tuned their PID controller using the Ziegler-Nichols technique. They began by obtaining a process reaction curve, running the system without PID control, and inputting a step change (6 km/h speed target). This curve highlighted the system's natural response properties. From this data, they extracted critical values: the maximum slope of the output-vs-time plot (S), the time delay (L), and the process gain (K). These values were then fed into Ziegler-Nichols formulas to calculate initial PID parameters. The proportional term (P) was implemented to provide the main control action and reduce rise time, the integral term (I) to eliminate steady-state error by summing the error over time, and the derivative term (D) to improve stability by responding to the error's rate of change. Following the implementation of these initial settings, the researchers fine-tuned the parameters iteratively to reach an ideal mix of rapid reaction, minimum overshoot, and steady-state accuracy that was specifically customized for their autonomous vehicle system. This tuning method produced a PID controller capable of keeping the vehicle's speed at the target speed while providing much better performance than the uncontrolled system.

The results showed considerable increases in the system's performance after the implementation of the PID controller. Without PID control, the system exhibited a large overshoot of 29\% and took 27 seconds to settle, with a steady-state error of 1.5 km/h. With the PID controller, the overshoot was reduced to just 4.9\%, the settling time decreased to 6 seconds, and the steady-state error was minimized to 0.02 km/h. The PID-controlled system demonstrated faster stability and better overall performance, successfully maintaining the vehicle's speed close to the desired target of 6 km/h with minimal fluctuations.

\subsection{Optimized self-adaptive PID speed control for autonomous vehicles}

The researchers in \cite{b7} conceived an adaptive PID control for autonomous vehicle speed control using two main approaches: Genetic Algorithm optimization (GA-PID) and Neural Network adaptation (NN-PID). In the Genetic Algorithm, the solution sets are called chromosomes.  A fitness function rates these chromosomes to determine how effective it is. The algorithm uses crossover and mutation operations to improve the solutions. There are multiple rounds of these operations also known as generations. Only fit chromosomes with high fitness scores are selected in each generation, while unfit chromosomes are discarded.

\begin{figure}[h]
\centerline{\includegraphics[width=0.4\textwidth]{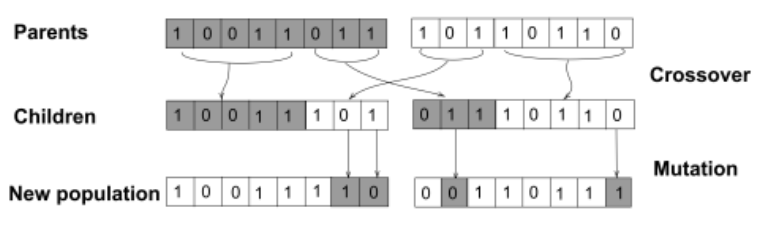}}
\caption{Genetic Algorithm}
\label{Genetic Algorithm}
\end{figure}

For PID tuning, the goal is to minimize the error between the current velocity and the desired velocity. To achieve this, the mean of squared error (MSE) is used as the objective function. For better results, the GA-PID algorithm optimization is performed for different scenarios, like varying speed profiles with and without external disturbances like wind speed and road slope.

\begin{figure}[h]
\centerline{\includegraphics[width=0.4\textwidth]{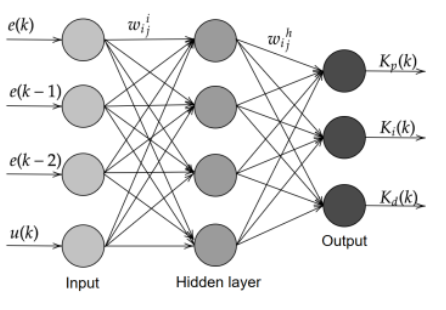}}
\caption{Neural Network}
\label{Neural Network}
\end{figure}

The NN-PID approach uses a feedforward neural network to adapt the PID gains online as compared to the offline GA algorithm.  The architecture of the NN consists of an input layer with four inputs, a hidden layer with four neurons and an output layer with three neurons corresponding to the PID gains as depicted in the figure \ref{Neural Network}.

They used the sigmoid activation function for the hidden layer and, interestingly, chose the ReLU function for the output layer to avoid predicting negative gains, which differs from some existing approaches. The neural network is trained using backpropagation of the error generated from previous PID gains. This allows for continuous adaptation of the controller to changing conditions and disturbances. The authors experimented with different numbers of hidden neurons and learning rates to optimize the NN-PID performance. 

The results of the study show that both GA-PID and NN-PID approaches provide effective adaptive control with good performance and disturbance rejection. The GA-PID controller often delivered smoother and quicker responses, particularly in disturbance-free scenarios. On the other hand, the NN-PID controller demonstrated better adaptability and robustness to disturbances, reacting quickly to error changes at each iteration. The GA-PID optimized for the scenario with disturbances like varying wind speed and road slope showed the best overall performance, with the lowest mean squared error (MSE) of 7.502, a rise time of 0.348 seconds, a settling time of 0.444 seconds, and an overshoot of 4.79\%. However, the NN-PID with 10 hidden neurons and a learning rate of 0.01 outperformed all other neural network configurations in terms of generalization across situations. The study concludes that although GA-PID may reach optimal performance in select conditions, NN-PID is more adaptable in a variety of real-world situations.

\subsection{Adaptive PID Control Design for Longitudinal Velocity Control of Autonomous Vehicles}

This authors of \cite{b8} present an adaptive PID (APID) control approach for longitudinal velocity control of autonomous vehicles. The authors propose a model reference adaptive PID controller to handle the nonlinear and time-varying dynamics of vehicle longitudinal motion. The controller adapts the PID gains to match the vehicle's behavior to a predefined reference model. The control law is given by:

\[u_{th} = k_p(v_d(t)-v(t)) - k_da(t) + k_i\]

where \(v_d\) is the desired velocity, v is the actual velocity, and a is the acceleration. The adaptation laws for the PID gains are derived using Lyapunov stability theory and are given by: 

\begin{align*}    
&\dot{k}_p = -\zeta_p\phi(t)(v_d - v(t))\\
&\dot{k}_d = -\zeta_d\phi(t)a(t)\\
&\dot{k}_i = \zeta_i\phi(t)
\end{align*}

where \(\zeta_p\), \(\zeta_d\), and \(\zeta_i\) are positive constants that adjust the convergence rate, and \(\phi(t)\) is a function of the error between the actual and reference model states. To improve robustness, the adaptation laws were modified with a "leakage" term to prevent unbounded growth of parameters in the presence of disturbances and changing operating conditions. This modification, known as \(\sigma\)-modification, adds a damping term when the parameter exceeds a predefined bound, e.g.,

\begin{multline*}
\dot{k}_p =
\begin{cases}
-\zeta_p\phi(t)(v_d - v(t)) & |k_p(t)| < \mathcal{K}_p\\
-\zeta_p\phi(t)(v_d - v(t)) - \sigma k_p(t) & |k_p(t)| \geq \mathcal{K}_p
\end{cases}
\end{multline*}
where \(\sigma\) and \(K_p\) are the constant parameters to be chosen.

The controller is designed to regulate both the throttle and brake actuators in order to maintain desired velocities ranging from 0 to 100 km/h. Two adaptive PID controllers have been implemented, one for throttle control and one for brake control, with a switching logic that selects the appropriate controller based on the need. The effectiveness of the approach is demonstrated through simulations in CarSim, a high-fidelity vehicle dynamics simulator. The results indicate strong velocity tracking ability under realistic settings such as road elevation changes, bends, and aerodynamic forces. The controller's adaptive nature enables it to manage different vehicle types and changing situations without requiring any modification. 

\section{Conclusion}

According to the findings of this survey, the PID for longitudinal control and Pure Pursuit for lateral control provide a reliable and efficient solution for autonomous vehicle navigation. These algorithms leverage well-established control theory principles to provide reliable and effective real-time vehicle motion control. PID controllers have proven effective at regulating vehicle speed and maintaining safe following distances, while Pure Pursuit algorithms enable autonomous vehicles to accurately track complex trajectories and navigate turns. Researchers have explored parameter tuning, novel error metrics, adaptive control strategies, and other types of optimization techniques to enhance the precision and responsiveness of both the algorithms for different and challenging real-world driving scenarios.

As the field of autonomous vehicles continues to advance, emerging control techniques such as stanley controller, model predictive control, deep reinforcement learning, and sensor fusion are beginning to complement the foundational PID and Pure Pursuit approaches. These newer algorithms aim to improve overall vehicle handling, enhance safety, and enable more complex autonomous maneuvers. While the core PID and Pure Pursuit methods remain integral components of autonomous vehicle control systems, the future of this field is likely to feature an increasingly diverse and sophisticated suite of control algorithms tailored to the growing capabilities and requirements of self-driving technology.

\section{Acknowledgment}
We extend our heartfelt gratitude to our parents, teachers, and friends for their unwavering support, with a special thanks to Dr. Bhakti Palkar for her exceptional guidance, which inspired us to explore this topic. Her insightful advice throughout this research has been an invaluable asset in shaping our study.

\end{document}